\documentclass{article}
\usepackage{fullpage}
\usepackage{graphicx} 
\usepackage{mathtools, amsfonts, amsthm}
\usepackage[svgnames]{xcolor}
\usepackage{natbib}
\usepackage{hyperref}
\usepackage{booktabs}
\usepackage{cleveref}
\usepackage{nicefrac}
\usepackage{xurl} 

\title{AI-rithmetic\thanks{Authors are listed in alphabetical order. Correspondence to \texttt{\{alexbie, tdick, kulesza, pragh, vinodraman, sergeiv\}@google.com}.}}

\author{
\begin{tabular}{c}
Alex Bie  \qquad Travis Dick \qquad Alex Kulesza \\
\vspace{-8pt}\\
Prabhakar Raghavan \qquad  Vinod Raman  \qquad Sergei Vassilvitskii \\
\\
\Large{Google}
\end{tabular}
}
\date{}

\usepackage{xcolor}
\definecolor{darkblue}{rgb}{0.1, 0.1, 0.45}
\hypersetup{
    colorlinks=true,
    linkcolor=darkblue,
    citecolor=darkblue,
    urlcolor=darkblue,
    filecolor=darkblue
}

\newif\ifshowcomments
\showcommentstrue 
\ifshowcomments
    \newcommand{\td}[1]{\textcolor{DarkGreen}{[Travis: #1]}}
    \newcommand{\ak}[1]{\textcolor{Purple}{[Alex K: #1]}}
    \newcommand{\ab}[1]{\textcolor{Orange}{[Alex B: #1]}}
    \newcommand{\vr}[1]{\textcolor{Blue}{[Vinod: #1]}}
    \newcommand{\sv}[1]{\textcolor{Red}{[Sergei: #1]}}
\else
    \newcommand{\td}[1]{}
    \newcommand{\ak}[1]{}
    \newcommand{\ab}[1]{}
    \newcommand{\vr}[1]{}
    \newcommand{\sv}[1]{}
\fi

\newcommand{\model}[1]{\texttt{#1}}

\begin{document}

\maketitle

\begin{abstract}
    Modern AI systems have been successfully deployed to win medals at international math competitions, assist with research workflows, and  prove novel technical lemmas. However, despite their progress at advanced levels of mathematics, they remain stubbornly bad at basic arithmetic, consistently failing on the simple task of adding two numbers. We present a systematic investigation of this phenomenon. We demonstrate empirically that all frontier models suffer significantly degraded accuracy for integer addition as the number of digits increases. Furthermore, we show that most errors made by these models are highly interpretable and can be attributed to either operand misalignment or a failure to correctly carry; these two error classes explain 87.9\%, 62.9\%, and 92.4\% of Claude Opus 4.1, GPT-5, and Gemini 2.5 Pro errors, respectively. Finally, we show that misalignment errors are frequently related to tokenization, and that carrying errors appear largely as independent random failures.
\end{abstract}

\section{Introduction}

Dramatic improvements in AI systems have led to surprising new applications in many areas, including advanced mathematics. While computer-assisted proofs have been around since the 4-color theorem of \citet{FourColor}, there has been an explosion of progress in the recent years. Large language models from Google and OpenAI have won medals in international competitions, developing proofs to new problems in algebra, geometry, and analysis~\citep{IMO-Gemini, IMO-GPT}. They have been used in research mathematics to find new lemmas, or improve existing proofs ~\citep{EconPaper, ALphaEvolve, Evolve2, Tao}. And mathematicians of all levels of experience, from graduate students to Fields medals winners, are using them to solve open problems and advance the state of the art \citep{fountoulakis-colt, sellke2025learningcurvemonotonicitymaximumlikelihood}.

At the same time, it is widely known (and we confirm here) that these systems are remarkably unreliable at basic arithmetic, sometimes failing to correctly add even two- or three-digit numbers. Similar failures such as an inability to count the number of `r's in ``strawberry'' or solve simple equations \citep{Equations} have also been observed. Of course, this is not inherently contradictory: arithmetic is different from advanced mathematics in many ways. But for humans, these skills are both clearly related and strongly ordered, arithmetic being a simple but critical part of the foundation on which deeper mathematics is built. Therefore, it may seem surprising that AI systems can outperform experts at more ``advanced'' tasks but fail to keep pace with kindergartners in the ``simple'' ones.

Accepting that AI systems do not exhibit human-like patterns of performance, we might instead re-imagine them as algorithmic machines performing complex behaviors by composing simple ones (as in, for example, long addition). Indeed, this is the idea behind the current wave of \emph{reasoning models} trained to perform a series of intermediate operations to reach a final answer. But this compositional view also fails to describe arithmetic performance, which we find is not even monotonic: AI systems are sometimes better at adding longer numbers than shorter ones. Moreover, while bigger models generally perform better than smaller models, all current models make significant numbers of mistakes as the problem size grows, suggesting that scaling alone is not likely to solve the problem, even though only a finite number of simple behaviors need to be learned and orchestrated by the reasoner.

Arithmetic, then, is an interesting case study in how modern AI systems can confound intuitions. In this work, we systematically investigate model behavior on simple integer addition problems, characterizing mistakes and identifying patterns. We find that errors are generally explainable in intuitive terms, offering a view of AI systems that is neither strictly human-like nor purely compositional, but idiosyncratic and dependent on design choices such as tokenization and auto-regression.

While tool use or task-focused training can probably be used to address the specific arithmetic difficulties we discuss here, the larger point is that today's models still have considerable weaknesses that we cannot fully anticipate. Manually correcting individual deficiencies as they are discovered does not seem like a viable path toward reliable general-purpose systems (if that is the goal). Instead, our aim is to use arithmetic as a convenient probe to understand some of today's AI failures, and hopefully begin to illuminate larger problems that have yet to be solved.

\begin{figure}
    \centering
    \includegraphics[width=0.9\linewidth]{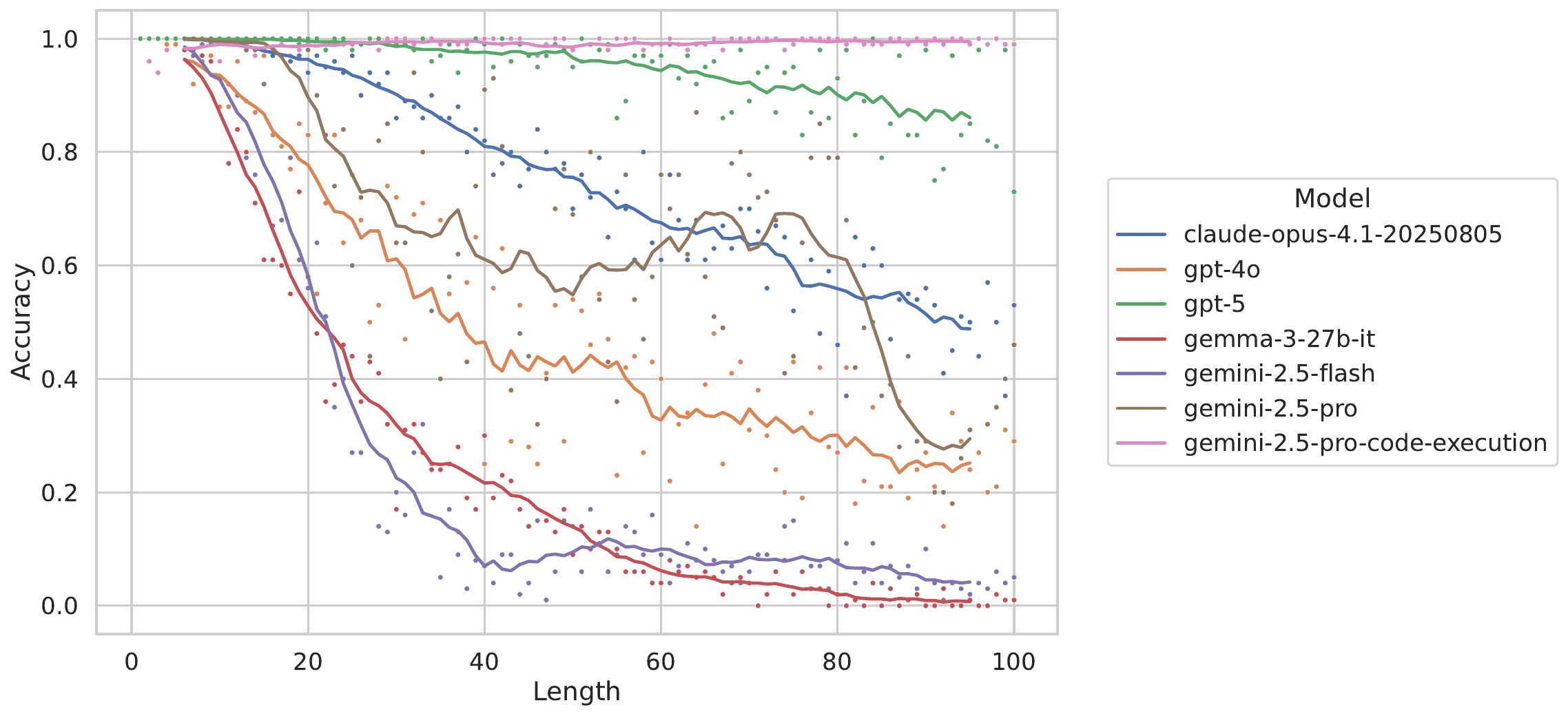}
    \caption{
    \textbf{Frontier models fail at adding long numbers.}
    The plotted points show the accuracy of the tested models on two-argument addition problems at each length. Since model performance fluctuates significantly, we include a centered 10-point moving average curve.}
    \label{fig:allModelsDegrade}
\end{figure}

\subsection{Summary of Findings}

We conduct systematic experiments on a collection of frontier large language models, finding that:
\begin{itemize}
    \item When adding two numbers, {\em all} frontier models perform poorly as the length of the numbers increases (Figure \ref{fig:allModelsDegrade}). We also find that performance is not necessarily monotone in length.
    \item Most observed mistakes are due either to {\em misalignment} or {\em close carry} errors (Figure \ref{fig:errorTypeByLength}). {\em Misalignment} errors are when one of the input operands is incorrectly shifted. {\em Close carry} errors are when the model incorrectly carries or fails to carry in borderline cases (that is, where the sum of digits is 9 but the model still carries a 1, or the sum is 10 and the model fails to carry). Misalignment and close carry errors comprise 87.9\%, 62.9\%, and 92.4\% of Claude Opus 4.1, GPT-5, and Gemini 2.5 Pro errors, respectively.
    \item Further investigation into error classes reveals that:
    \begin{itemize}
        \item Misalignment errors are often periodic with respect to argument length, a fact that may be explained by tokenization, especially for models that use multi-digit tokens (Section \ref{sec:tokenization}).
        \item The occurrence of close carry errors is largely consistent with an \emph{independent error model} where models err on each close carry independently with probability $p$ (Section \ref{sec:independence}). If an addition problem contains $n$ close carries, this predicts a $(1-p)^n$ chance of getting the problem correct (i.e., the \citet{lecun-doomed} model), and a geometrically distributed first error position.
    \end{itemize}
\end{itemize}

\subsection{Related Work}

Although language models can perform arithmetic tasks with non-trivial accuracy in certain settings \citep{wei2022emergent,yang2023gpt,maltoni2024arithmetic}, their failure to generalize and overall poor performance have been widely observed \citep{saxton2019analysing,nogueira2021investigating,dziri2023faith,qian2023limitations,gambardella2024language,testolin2024can,yan2025large, substack}.

Recent work has focused on understanding how AI systems represent numbers or implement arithmetic operations \citep{stolfo2023mechanistic,dziri2023faith,nikankin2025arithmetic,deng2024language,zhang2024interpreting,baeumel2025modular,levy2025language,kantamneni2025language}, as well as how they can be improved. Techniques include prompting or training with chain-of-thought \citep{wei2022chain,lee2023teaching}, augmenting the model's abilities with symbolic systems \citep{yang2024arithmetic,dugan2024occamllm}, adding hints or restructuring the problem \citep{nogueira2021investigating,zhou2023algorithms}, increasing numerical precision \citep{feng2024numerical}, and utilizing improved positional encodings \citep{shen2023positional,mcleish2024transformers,zhou2024transformers}.

In terms of error analysis, \citet{singh2024tokenization} find that enforcing right-to-left tokenization with separators improves accuracy for GPT-4 (which employs multi-digit tokenization), and that errors are highly \emph{position-dependent} rather than problem difficulty-dependent. In this vein, our analysis uncovers that accuracy is periodic with respect to length (Figure \ref{fig:gptPeriodic}) for GPT models; furthermore we identify that these tokenization-induced errors are due to misalignment.
\citet{baumel2025lookahead} analyze close carry errors, attributing them to the inability of models to anticipate cascading carries. They conduct an in-depth analysis on multi-operand, 3-digit addition on smaller models, while we examine two-operand, $n$-digit addition and frontier reasoning models. Our focus on operand length leads us to uncover the independent nature of close carry errors. \citet{nikankin2025arithmetic} use tools from mechanistic interpretability to identify the exact circuits that pretrained LLMs use to add, observing that an arithmetic problem activates several independent (flawed) mechanisms that contribute to promoting the correct answer. \citet{nikankin2025arithmetic} primarily study Llama models, and restrict their focus to answers computed in a single forward pass (e.g. input operands and the output answer are a single 3-digit token).

\section{Model Performance}

\begin{figure}
    \centering
    \includegraphics[width=0.9\linewidth]{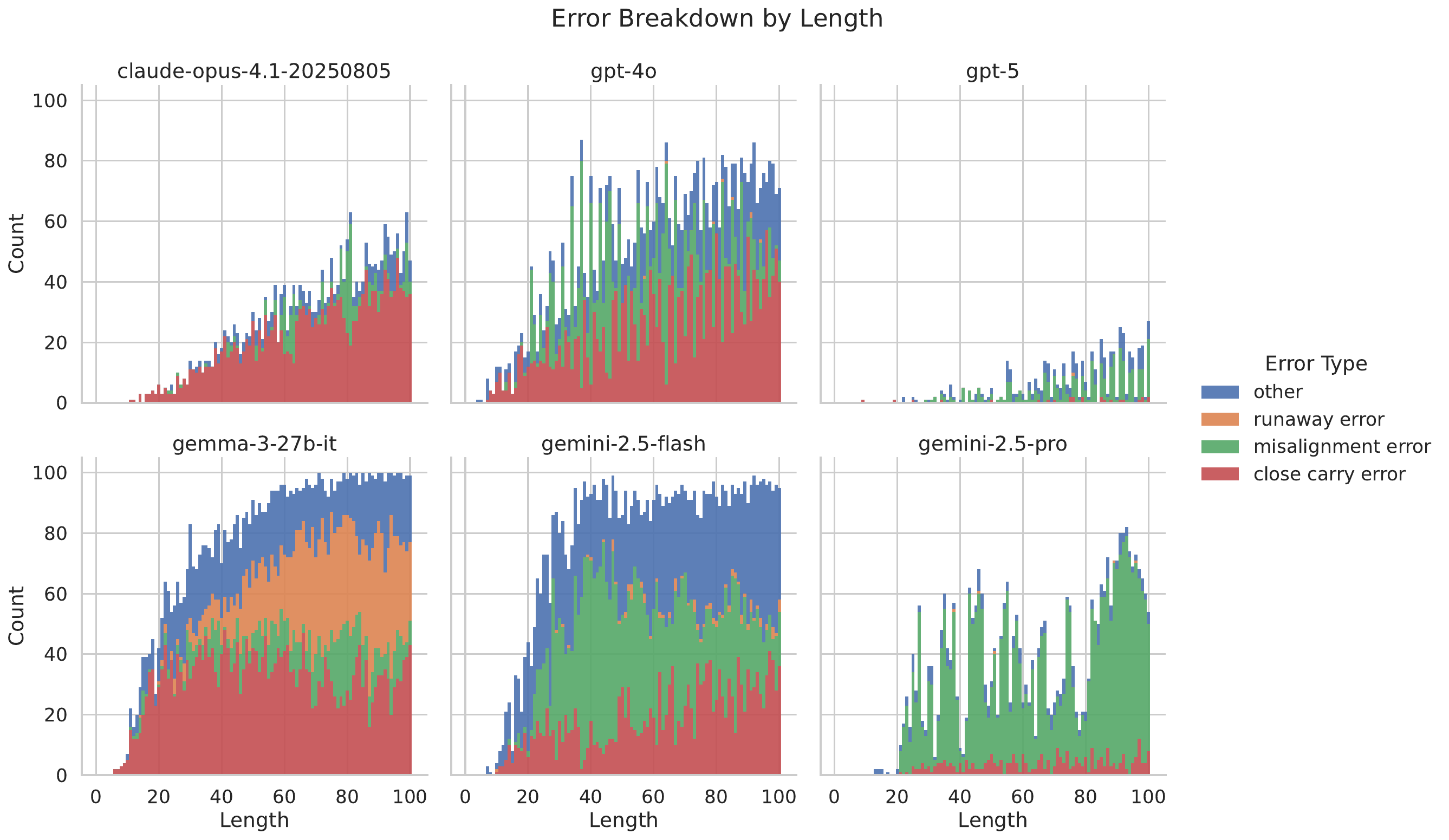}
    \caption{\textbf{Most errors can be explained}. We plot the distribution of error types at each length. For a description of error types, see Section \ref{sec:exploration}. We find that close carry and misalignment errors cover 87.9\% of Claude Opus 4.1's mistakes, 62.9\% of GPT-5's mistakes, 92.4\% of Gemini 2.5 Pro's mistakes, and at least 55.6\% of every model's mistakes.}
    \label{fig:errorTypeByLength}
\end{figure}

In this section, we investigate the performance of frontier LLMs on adding two numbers.

\paragraph{Setup.} We use prompts of the form:
$$\texttt{What is $A$ + $B$? Write just the answer.}\footnote{We also tested several other prompts, but did not find any that elicited significantly better accuracy. We argue that an intelligent system should not depend meaningfully on the form of the prompt as long as the intent is clear.}$$

For each length $d \in \{1,\ldots, 100\}$, we generate 100 prompts by setting
$A$ and $B$ to be $d$-digit numbers chosen independently and uniformly at random from $\{10^{d-1}, \ldots, 10^d-1\}$. We send each prompt to each model and collect their responses.

To extract an answer from the model's response, we remove all commas (to account for different numerical formatting conventions) and then extract the span of digits with the lowest edit distance to the correct sum $A+B$. We say that the model is correct if the match is exact, and incorrect otherwise. This test is lenient: if the model response includes multiple spans of digits, we consider the response correct if any of them is correct. \Cref{tab:correctIncorrectExamples} gives examples of several correct and incorrect model responses when the true sum is \texttt{1234}.

\begin{table}[h]
    \centering
    \begin{tabular}{lc}
        \toprule
        \textbf{Response} & \textbf{Correct / Incorrect}\\
        \midrule
        \texttt{1234} & \texttt{correct} \\
        \texttt{1,234} & \texttt{correct} \\
        \texttt{The answer is 1234.} & \texttt{correct} \\
            \texttt{Maybe the answer is 1234, but really it is 9999.} & \texttt{correct} \\
        \texttt{9991234000} & \texttt{incorrect} \\
        \texttt{1 234} & \texttt{incorrect} \\
        \texttt{5678} & \texttt{incorrect} \\
        \bottomrule
    \end{tabular}
    \caption{Examples of grading model responses when the true answer is \texttt{1234}.}
    \label{tab:correctIncorrectExamples}
\end{table}

\paragraph{Models.} We test a variety of frontier LLMs, including Claude Opus 4.1, GPT-4o, GPT-5, Gemini 2.5 Flash, Gemini 2.5 Pro, Gemini 2.5 Pro with Code Execution, and Gemma 3 27B. All models tested have been consumer facing at some point in their life cycle.

\paragraph{Results.} \Cref{fig:allModelsDegrade} shows the fraction of correct responses a function of the length $d$. With the exception of Gemini 2.5 Pro Code Execution, the accuracy of all models drops significantly below 100\% as the number of digits increases. Even at a moderate length of 20, most models exhibit a nontrivial error rate. This suggests that the varieties of architectures, data mixtures, and training practices used by current models are not sufficient for learning to accurately perform basic addition.

Gemini 2.5 Pro with Code Execution's stronger performance can be attributed to the fact that it usually calculates the sum using short python scripts. Nonetheless, we still observe errors. These occur when the model fails to use the tool or fails to correctly copy the answer calculated by the tool.

\section{Exploration of Mistakes}
\label{sec:exploration}

Next, we explore the types of mistakes evident in \Cref{fig:allModelsDegrade}.
We find that a significant fraction of mistakes can be explained intuitively.

Recall that when adding a pair of numbers, the canonical long addition algorithm aligns the operands vertically and then computes a sequence of digit sums, one for each column, proceeding from least to most significant position. When a column sum exceeds 9, a one is ``carried'' to the next (more significant) column.

A large proportion of model mistakes are consistent with one of two natural failure modes during the execution of this algorithm: (1) misalignment, where the digits of one operand are shifted by a few positions in either direction, and (2) close carry failure, where a carry is performed in error when the column sum is 9, or not performed when the column sum is 10. (See below for the details of how we identify such errors.)

\Cref{tab:errorExplanationRate} shows the proportion of all incorrect answers given by each model that are consistent with one of these two error types.
Notably, these two types cover 92.4\% of Gemini 2.5 Pro's mistakes, 87.9\% of Claude Opus 4.1's mistakes, and at least 55.6\% of every model's mistakes.

\begin{table}[h]
    \centering
    \begin{tabular}{rc}
    \toprule
    \textbf{Model} & \textbf{Proportion of errors explained} \\
    \midrule
    \model{claude-opus-4.1} & 87.9\%\\
    \model{gpt-4o} & 77.5\%\\
    \model{gpt-5} & 62.9\%\\
    \model{gemini-2.5-flash}  & 60.4\%\\
    \model{gemini-2.5-pro} & 92.4\%\\
    \model{gemma-3-27b-it} & 55.6\%\\
    \bottomrule
    \end{tabular}
    \caption{The fraction of all mistakes across 10,000 sum problems that are either misalignment or close carry errors.}
    \label{tab:errorExplanationRate}
\end{table}

\paragraph{Misalignment Errors.}

We say that a model's extracted answer for the sum of $A$ and $B$ contains a \emph{misalignment error} if at least the first 6 digits of the extracted answer match the sum when adding $A$ and $B$ with the arguments offset by up to 10 digits in either direction. We limit the test to six digits because models tend to compound their errors as they go, making long misaligned matches unlikely. We require at least six digits to avoid false detections. 

More specifically, for each offset $s \in \{-10, \ldots, -1, 1, \ldots, 10\}$, we calculate $\operatorname{offsetsum}(A, B, s) = A \cdot 10^{\max(0, s)} + B \cdot 10^{\max(0, -s)}$ and calculate the length of the shared prefix between the extracted model answer and $\operatorname{offsetsum}(A,B,s)$. We identify the response as a misalignment error if the length is at least 6 and the extracted answer is not equal to one of the arguments.\footnote{When the model outputs $A$ or $B$, we do not consider this to be a misalignment error, even though $\operatorname{offsetsum}(A,B,s)$ typically begins with $|s|$ digits from $A$ or $B$.} Note that matching 6 digits for an offset of size at most 10 is unlikely by chance: there are at most 20 distinct 6-digit prefixes obtained from the offset sums, yet approximately $10^6$ possible prefixes, so the probability of a uniformly drawn answer matching a six-digit prefix is $\approx 0.002\%$. 

\Cref{tab:misalignmentErrorExample} shows examples of the prefixes that qualify as misalignment errors for $A = 555555$ and $B = 123456$.

\begin{table}
    \centering
    \begin{tabular}{crrrc}
    \toprule
    \textbf{Offset $s$} 
    & \textbf{$A \cdot 10^{\max(0,s)}$}
    & \textbf{$B \cdot 10^{\max(0,-s)}$}
    & \textbf{$\operatorname{offsetsum}(A,B,s)$}
    & \textbf{Required response prefix}\\
    \midrule
    -2 & 555555 & 12345600 & 12901155 & \texttt{129011} \\
    -1 & 555555 & 1243560 & 1790115 & \texttt{179011} \\
    1 & 5555550 & 123456 & 5679006 & \texttt{567900} \\
    2 & 55555500 & 123456 & 55678956 & \texttt{556789} \\
    \bottomrule
    \end{tabular}
    \caption{A collection of 6-digit response prefixes that would count as a misalignment error for the sum of $A = 555555$ and $B = 123456$.
    Each row corresponds to an offset $s$ and includes the offset arguments, the offset sum, and the prefix a model answer must have to be considered a misalignment mistake.
    We show offsets of up to 2 digits in either direction, but our implemented test searches for offsets of up to 10 digits.}
    \label{tab:misalignmentErrorExample}
\end{table}

\begin{table}
    \centering
    \begin{tabular}{rc}
    \toprule
    \textbf{Problem} & \textbf{Possible close carry errors} \\
    \midrule
    \texttt{10+19 = 29} & \texttt{3*} are close carries but nothing else.\\
    \texttt{11+19 = 30} & \texttt{2*} are close carries but nothing else.\\
    \texttt{10+10 = 20} & No close carries, despite 0s in answer (the rightmost column sum is 0).\\
    \texttt{199+199 = 398} & No close carries, despite 9s in answer (the middle column sum is 19).\\
    \bottomrule
    \end{tabular}
    \caption{Several addition problems and a description of the set of responses that would count as a close carry error. The character \texttt{*} is a placeholder that can be any digit.}
    \label{tab:closeCarryExamples}
\end{table}

\paragraph{Close Carry Errors.}
We use the following procedure to identify close carry errors.
If the correct sum and the extracted model answer are different lengths, we pad the shorter to the left with zeros.
Then we compare the extracted answer to the correct answer digit by digit from left to right---although long addition is performed from right to left, LLMs generate tokens autoregressively in the order they appear.
When we reach the first digit at which an error occurs, we check if it is consistent with a close carry error:
\begin{itemize}
    \item If the digit is too large by 1 and the column to the right has a correct column sum of 9, then we say it is a close carry error. 
    \item If the digit is too small by 1 and the column to the right has a correct column sum of 10, then it is also a close carry error.
\end{itemize}
If the response contains no errors, or the first error does not satisfy either case above, then we do not identify it as a close carry error.
\Cref{tab:closeCarryExamples} show some examples of the errors that would be considered close carries for small addition problems.

\paragraph{Runaway errors.} In addition to misalignment and close carry errors, we observed that some of the models frequently output answers that are much longer than the correct answer (sometimes thousands of digits long).
We say that an extracted answer is a \emph{runaway} error if its length is at least 50\% longer than the true answer.

\begin{figure}
    \centering
    \includegraphics[width=0.9\linewidth]{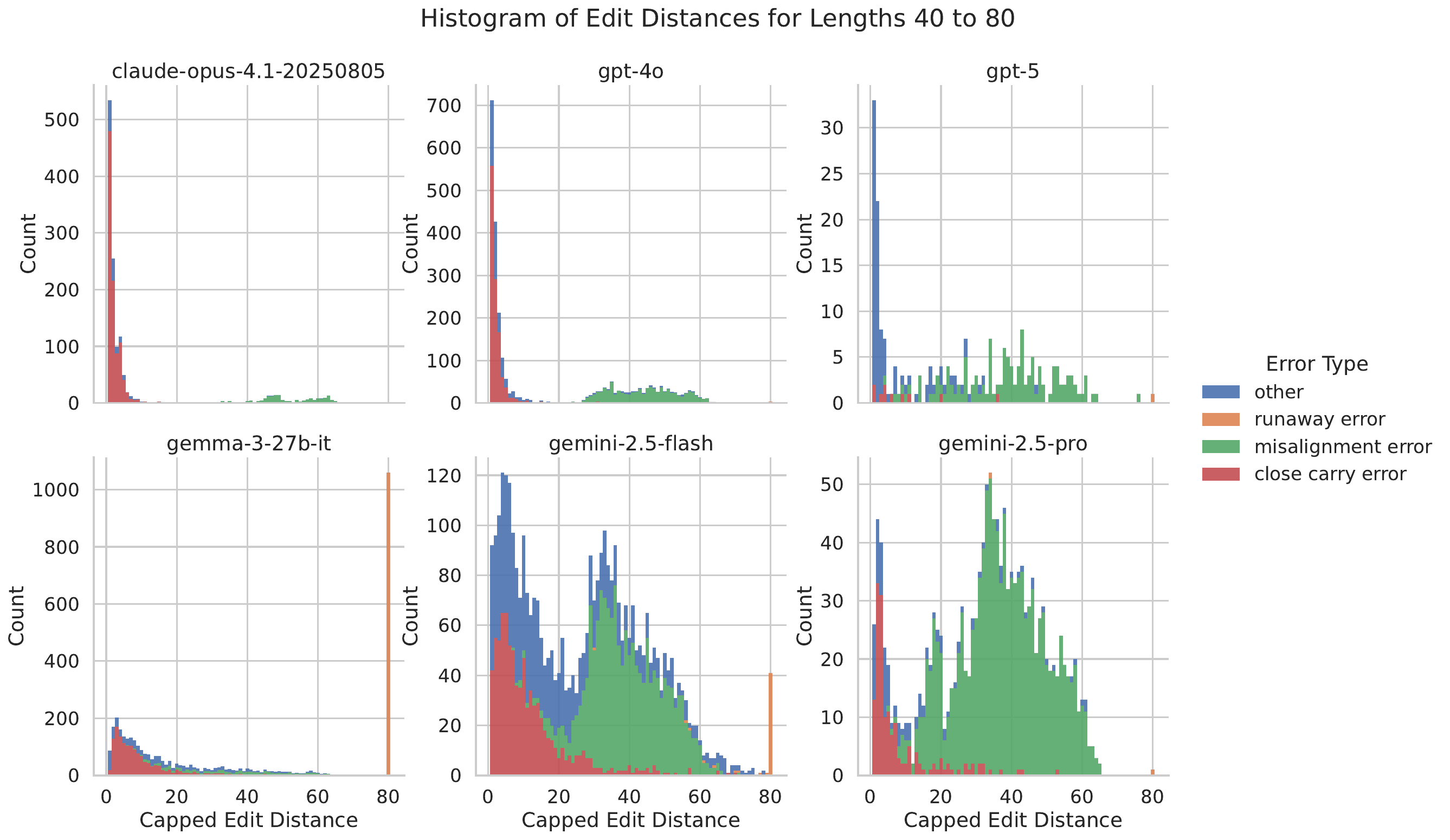}
    \caption{\textbf{Bimodal edit distance distribution of mistakes corresponds to error classes}. We plot a histogram of edit distance from the true answer for all mistakes, and color by its error classification. We observe a spike of small edit distance errors for close carry, and a spike of large edit distance errors for misalignment.}
    \label{fig:editDistanceHistogram}
\end{figure}

\paragraph{Results.}

\Cref{fig:errorTypeByLength} shows how the mistakes made by each model break down into aforementioned mistake types. Since the error types we consider are not mutually exclusive, we classify a model's response according to the first mistake type it matches in the following order: runaway, misalignment, close carry.\footnote{The overlap between error types is low, and the order here does not have a large impact on the results. Among all models, we observe 8081 responses that are only identified as misalignment errors, 9347 that are only identified as close carries, and 543 that are identified as both.
}

Any mistakes that do not match the conditions for these three error types are classified as ``other''.
For Claude Opus 4.1, Gemini 2.5 Pro, GPT-4o, and GPT-5, the majority of their mistakes match at least one of the three mistake types.
Most of the mistakes made by Claude Opus 4.1 are close carrying errors, while most of the mistakes made by Gemini 2.5 Pro and GPT-5 are misalignment errors.

\Cref{fig:editDistanceHistogram} shows histograms of the edit distance between the extracted model answer and the true answer for mistakes on problems of length 40 to 80.
Each histogram bar is colored according to the types of the mistakes.
Most models exhibit a bimodal distribution with a significant fraction of the low edit distance mistakes classified as close carry errors and many of the longer edit distance mistakes classified as misalignment errors. This is intuitive since close carry errors are local in nature, while a single misalignment can distort the entire result.

\begin{figure}
    \centering
    \includegraphics[width=0.8\linewidth]{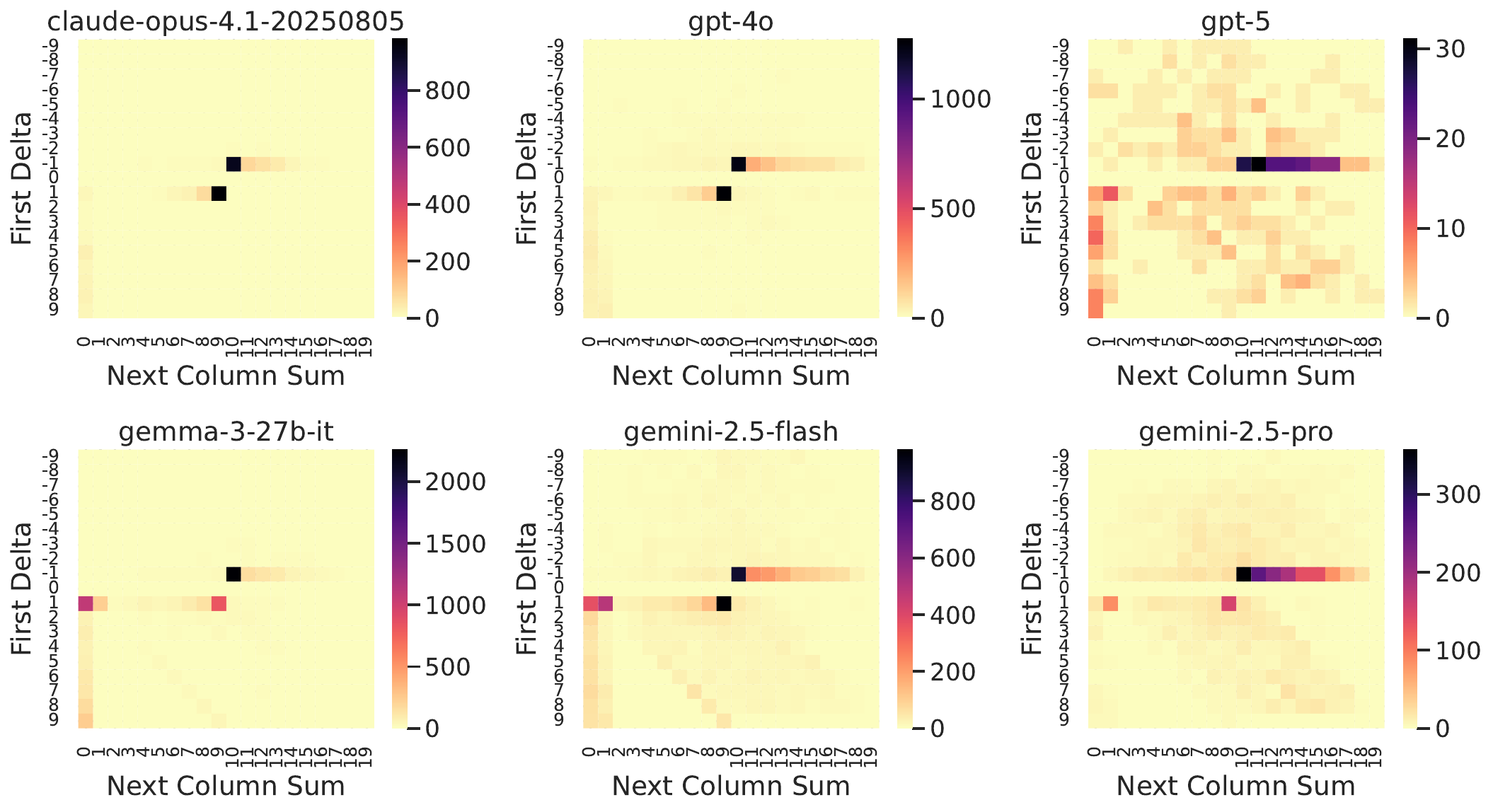}
    \caption{For each incorrect response made by a model, we find the left-most incorrect digit and calculate (a) the delta between it and the correct digit, and (b) the long addition column sum one position to the right (i.e., the column that would carry into the incorrect column).
    This figure shows how common each digit delta and next column sum are for each model.
    A significant fraction of the total count falls on the (delta=-1, next sum=10) and (delta=1, next sum = 9) positions, which exactly characterizes close carry mistakes.
    }
    \label{fig:deltaVsColumSum}
\end{figure}

\Cref{fig:deltaVsColumSum} shows that close carry errors are significantly more prevalent than other carrying errors. The heat maps show the frequency of the model's leftmost error delta (the difference between model's digit and answer's digit) and the column sum in the next position. Close carry errors are characterized by having a delta of -1 and next column sum of 10, or a delta of +1 and a next column sum of 9.
We see that these combinations are much more frequent than any other combination, although each model has its own unique behaviors as well.

\begin{figure}
    \centering
\includegraphics[width=0.75\linewidth]{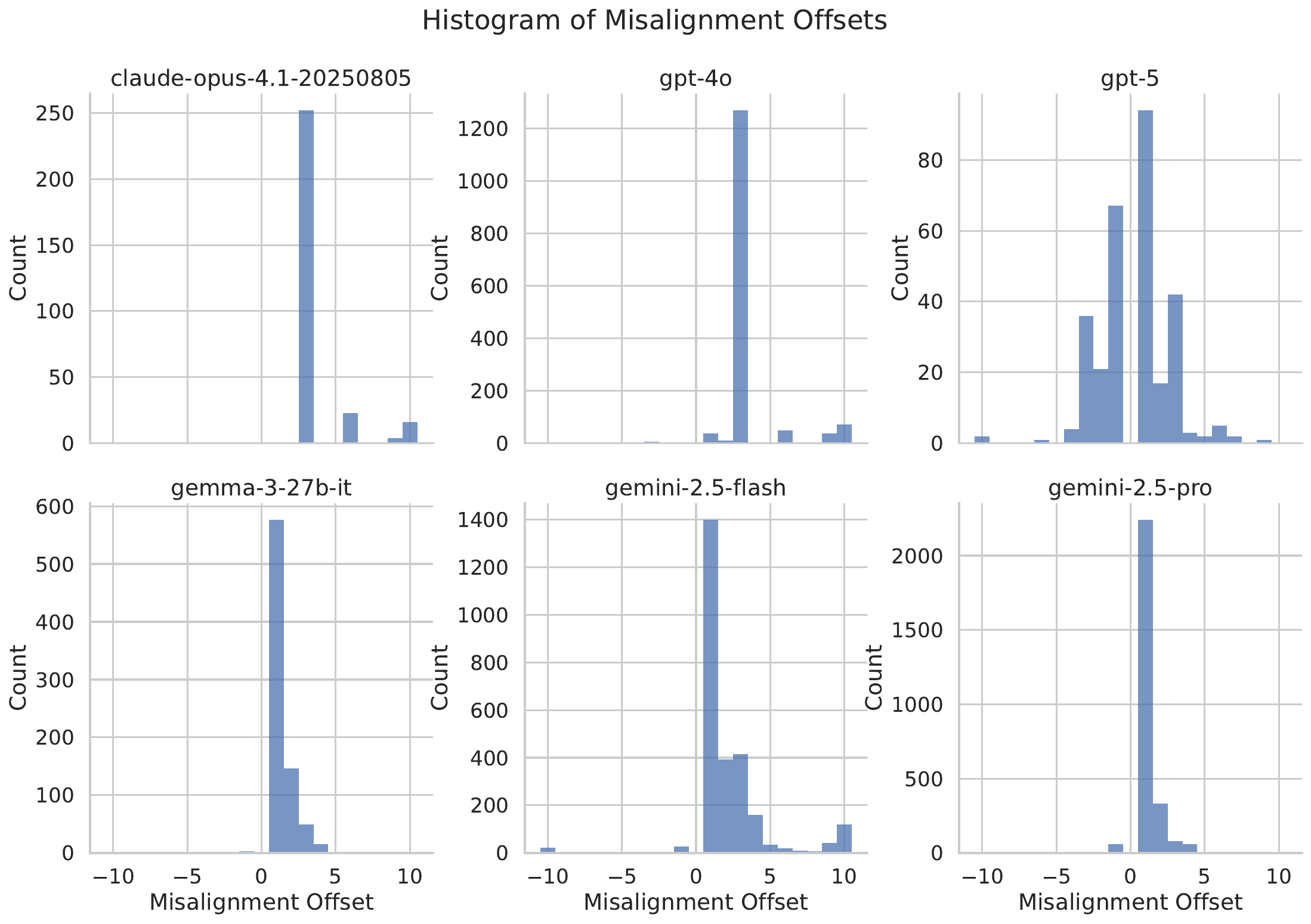}
    \caption{The misalignment offset that produces the longest misaligned prefix match on examples classified as a misalignment error. Models are biased towards positive offsets (corresponding to a rightward shift of the second argument). Note the modal offset of 3 for Claude Opus 4.1 and GPT-4o, and 1 for Gemini and Gemma; we attribute this to digit tokenization (see Section \ref{sec:tokenization} for further discussion).}
    \label{fig:misalignmentOffsets}
\end{figure}

Finally, for misalignment errors, \Cref{fig:misalignmentOffsets} shows histograms of the offsets producing the longest prefix match.
Recall that positive offsets correspond to padding the first argument on the right with 0s.
With the exception of GPT-5, the models all nearly always have a positive offset.
A possible explanation is that it is common practice to write the longer argument first when adding numbers of different lengths.
We see that Gemini 2.5 Flash and Pro, and Gemma 3 27B all typically misalign the arguments by a single digit position.
On the other hand, GPT-4o and Claude Opus 4.1 favor a $3$-digit misalignment. Note that these plots also show some evidence of periodicity in the offset frequencies for some models; we will return to this observation in \Cref{sec:tokenization}.

\section{Explanation of Mistakes}

\subsection{Tokenization and Misalignment}
\label{sec:tokenization}

As observed in \Cref{fig:misalignmentOffsets}, misalignment errors tend to favor certain argument offsets, which vary by model. Here we investigate further and present evidence that misalignment errors may be related to tokenization.

\begin{figure}
    \centering
    \includegraphics[width=0.45\linewidth]{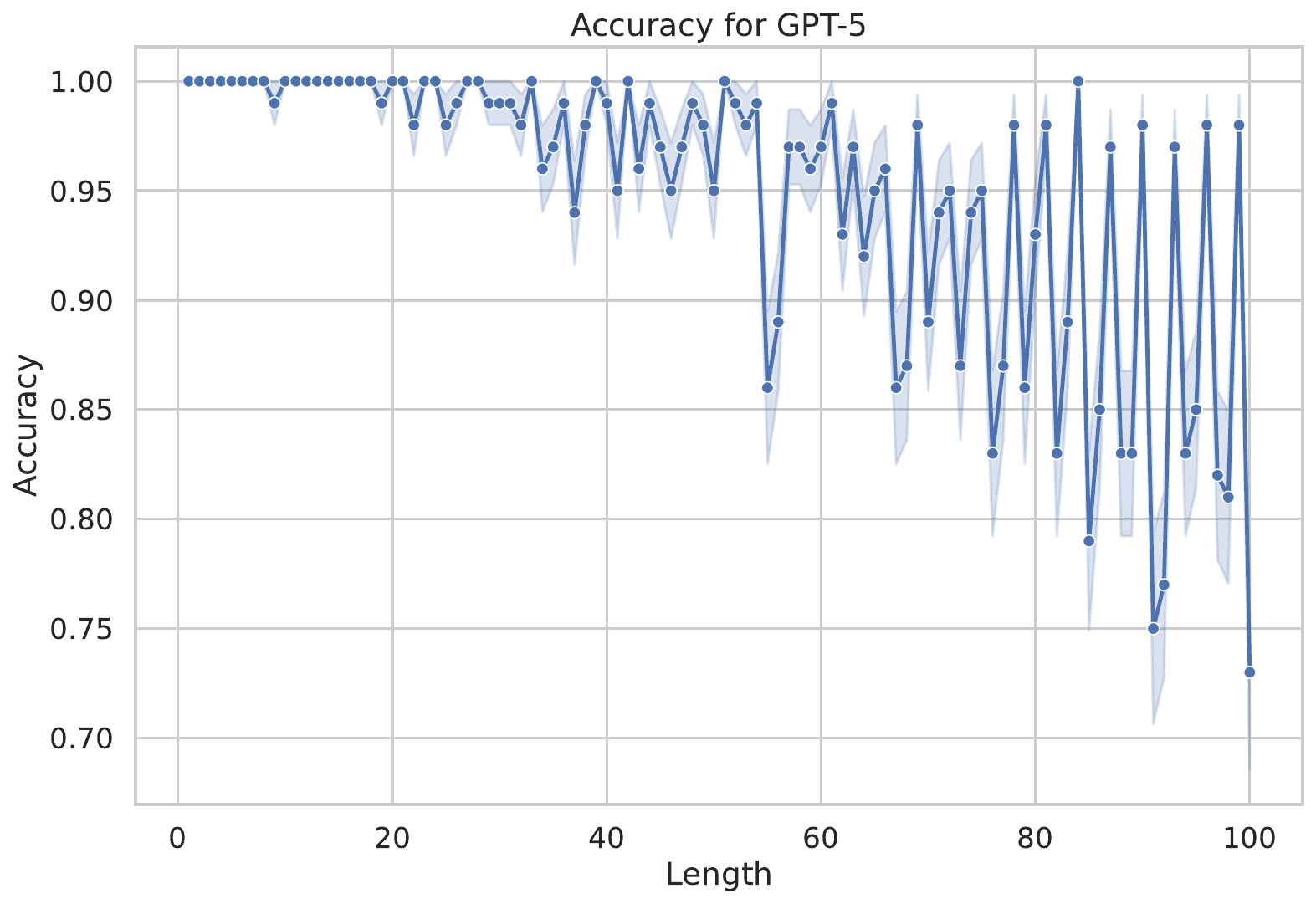}
    \qquad
    \includegraphics[width=0.45\linewidth]{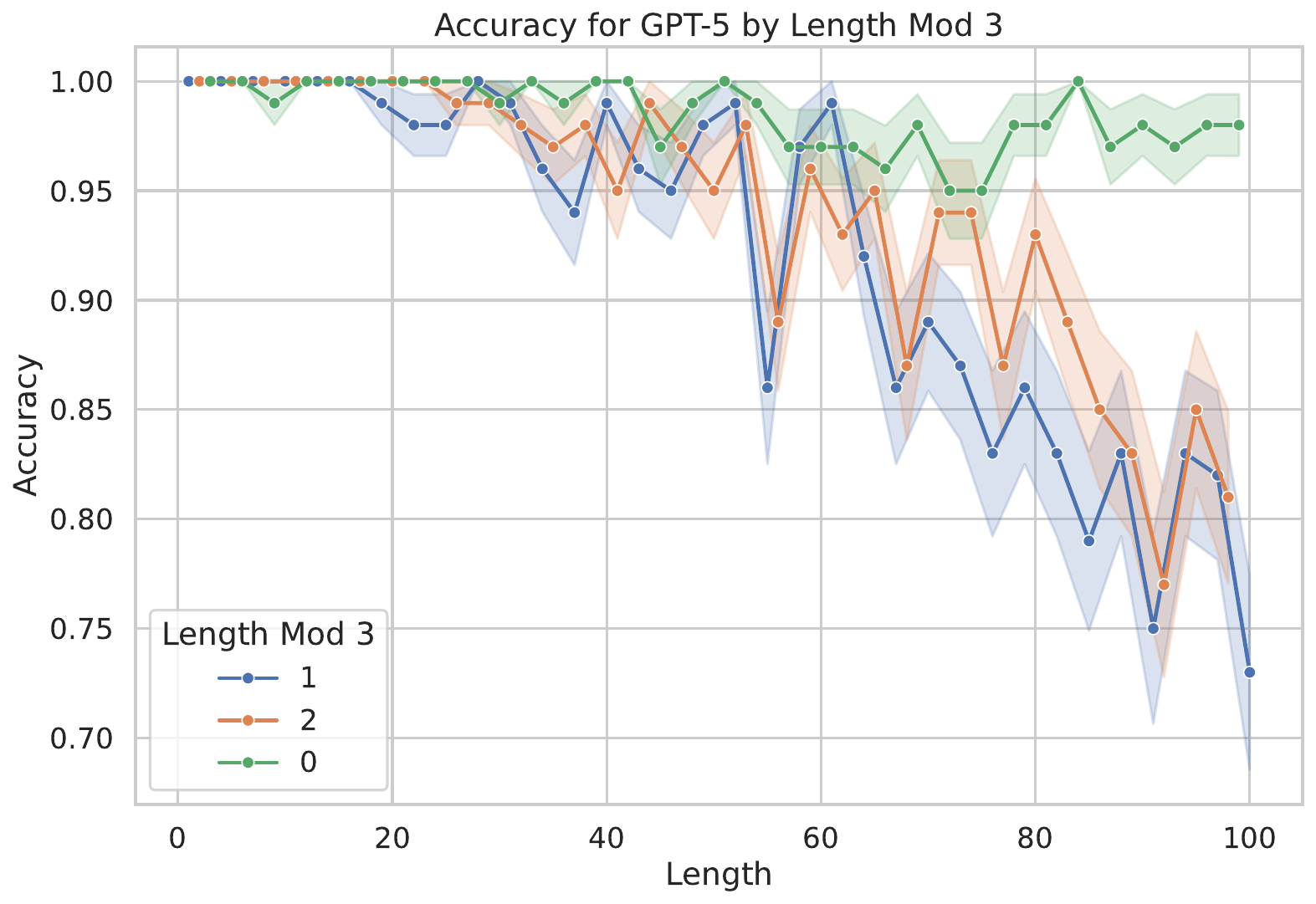}
    \caption{\textbf{Left:} Plot showing the accuracy of GPT-5 as a function of length. Notice that every third accuracy is higher than the rest. \textbf{Right:} Plotting the same data but split into three curves, where each curve contains lengths that are equivalent to 0, 1, or 2 mod 3. The shaded region shows the standard error of the accuracy estimate, indicating that the separation of the 0 curve is significant.}
    \label{fig:gptPeriodic}
\end{figure}

\Cref{fig:gptPeriodic} shows the unsmoothed accuracy of GPT-5 (which makes primarily misalignment errors) as a function of the argument length.
Particularly for longer arguments, accuracy seems to behave in a periodic way.
The right plot splits this curve into three parts, corresponding to lengths that are equivalent to 0, 1, or 2 mod 3.
With the shaded regions indicating standard error, it is apparent that the accuracy of GPT-5 is significantly higher for arguments whose length is a multiple of three. 

GPT-5 uses the \texttt{tiktoken} tokenizer \citep{tiktoken, gpt-tokenizer}, in which a pre-tokenization regular expression breaks spans of digits into groups of 3 digits, followed by a group containing the remaining 1 or 2 digits if the span length is not a multiple of three. 
Each of these groups is then represented by a single token.
As a consequence, the tokenization of a $d$-digit number begins with $\lfloor d/3\rfloor$ tokens that each represent 3-digit strings, followed by a token representing a 1- or 2-digit string if $d \mod 3 \neq 0$. \Cref{fig:gptPeriodic} therefore suggests that GPT-5 is more likely to make mistakes precisely when 1- or 2-digit tokens are present.

\begin{figure}
    \centering
    \includegraphics[width=0.85\linewidth]{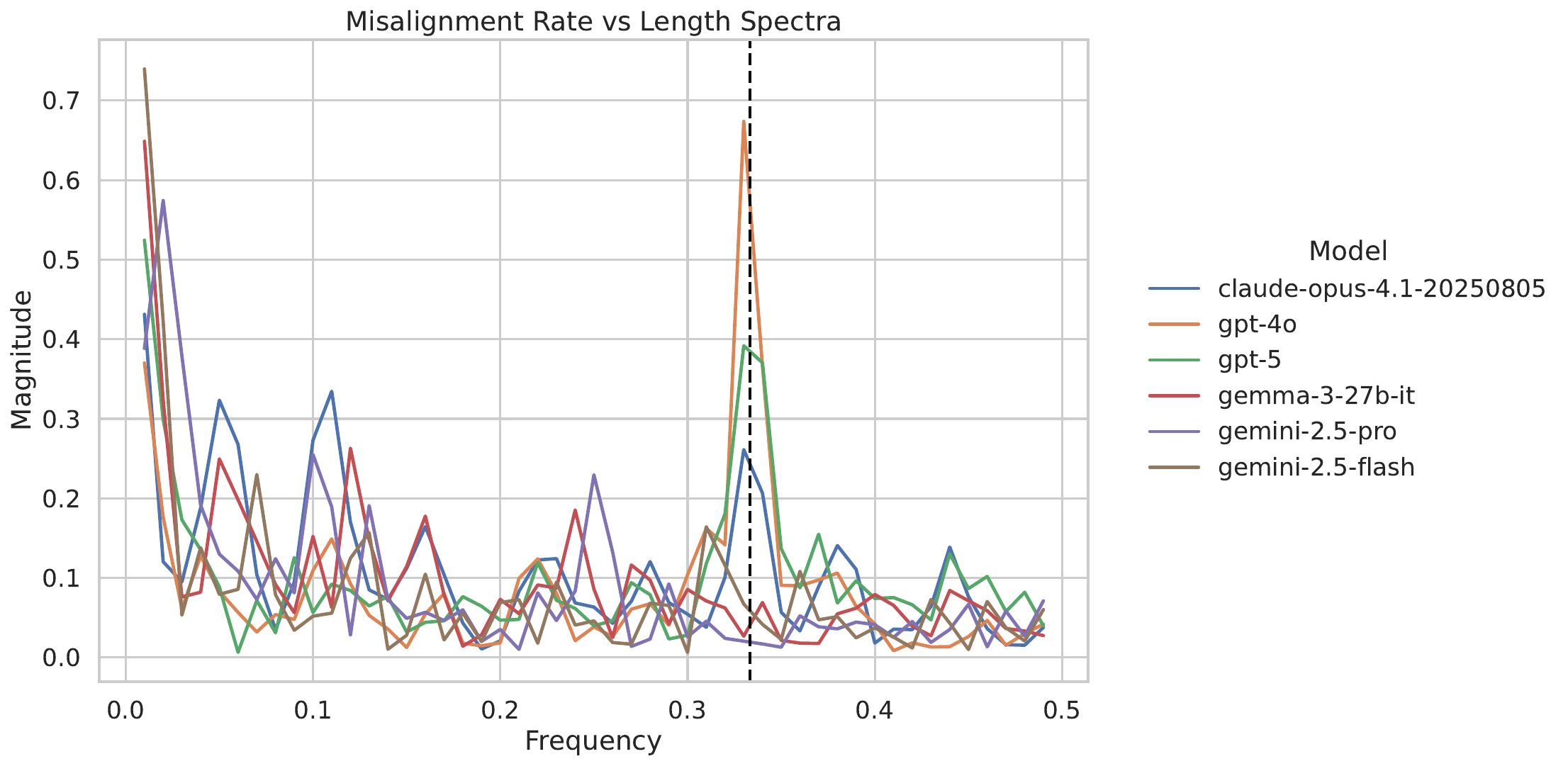}
    \caption{Spectra of the misalignment rate vs. length signals for each model. Each spectrum has been normalized to have unit $\ell_2$ norm. The dashed vertical line indicates a frequency of $\nicefrac{1}{3}$.}
    \label{fig:fftMisalignmentRate}
\end{figure}

To investigate whether this phenomenon explains misalignment errors more generally, we plot in \Cref{fig:fftMisalignmentRate} the discrete Fourier transform of the 
misalignment error rate as a function of length for all of the models. As expected, we see a large magnitude for the frequency $\nicefrac{1}{3}$ for GPT-5, corresponding to the reduced error rate on lengths divisible by 3. In addition, we see pronounced spikes at $\nicefrac{1}{3}$ for GPT-4o, which also uses a 3-digit tokenizer \citep{gpt-tokenizer}, and Claude Opus 4.1, for which we find strong evidence of using a similar tokenization scheme.\footnote{Specifically, when querying the \texttt{count\_tokens} API for \texttt{claude-opus-4-1-20250805} \citep{claude-4-1-tokenizer}, the increase in returned token count as a function of the number of digits $d$ is precisely $\lceil d/3 \rceil$, matching GPT-5 and GPT-4o.} Meanwhile, we see no spikes at $\nicefrac{1}{3}$ for the other models, for which we find strong evidence that they tokenize digits individually.\footnote{Gemma 3's tokenizer tokenizes digits individually \citep{gemma-tokenizer}. For evidence that Gemini 2.5 tokenizes digits individually: \cite{gemmateam2025gemma3technicalreport} reports using the same tokenizer as Gemini 2; and the \texttt{count\_tokens} API for Gemini 2.5 Pro and Flash reports an increase of $d$ tokens when passed a $d$-digit number.}

These results suggest that misalignment errors may be exacerbated when operands are encoded with tokens containing variable numbers of digits. However, we also note that models using single-digit tokenizers tended to exhibit more misalignment errors overall, so it may be that longer tokens actually help to \emph{reduce} misalignment errors, and this effect is muted when mixed-length tokens are used.

\subsection{Independence and Close Carry Errors}
\label{sec:independence}

\begin{figure}
    \centering
    \includegraphics[width=0.92\linewidth]{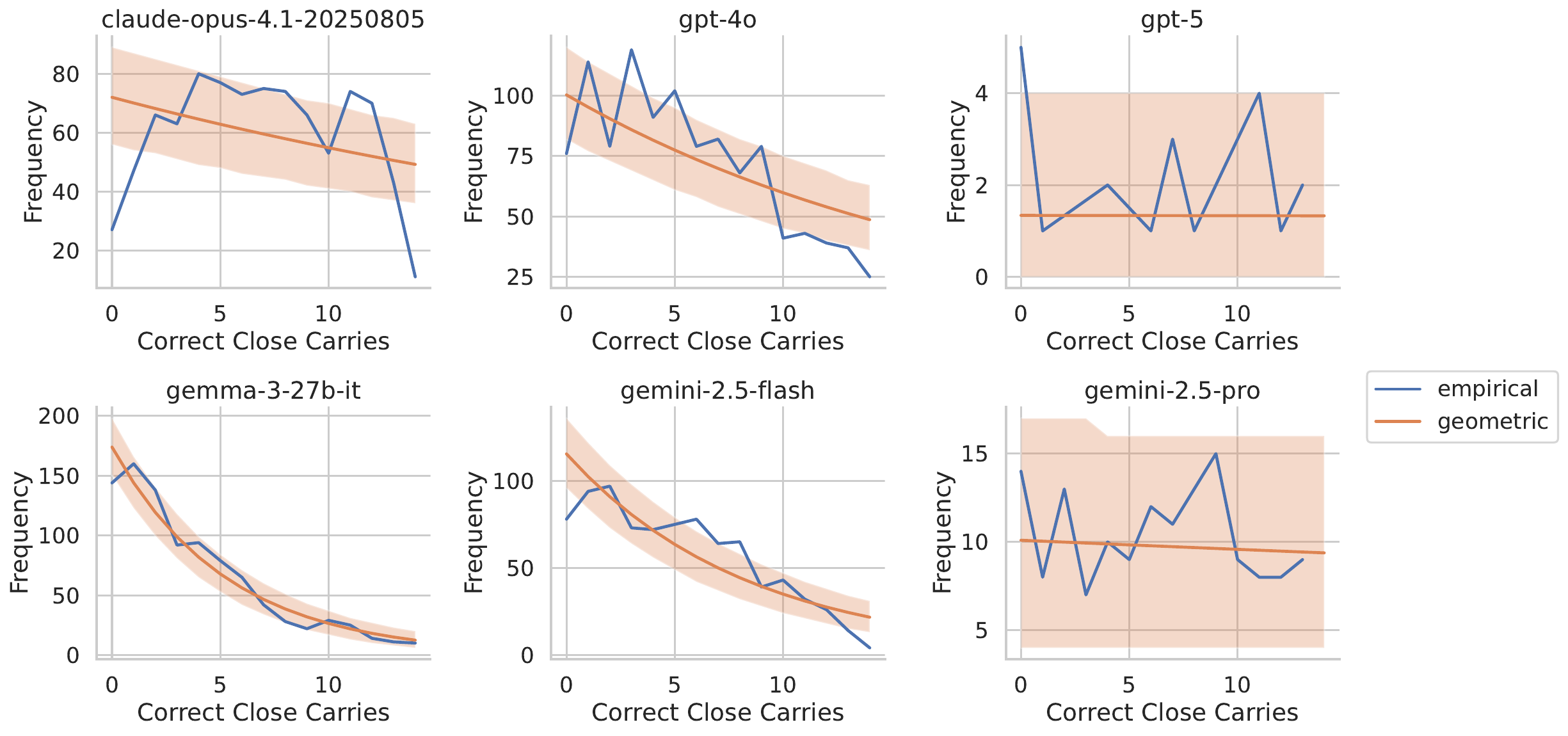}
    \caption{The number of correct close carries observed prior to the first close carry error. We compare to a simple geometric model that assumes the LLM succeeds on each close carry independently with probability $p$; this model predicts exactly $i$ correct carries with probability $p(1-p)^{i}$. The shaded region shows the 95\% confidence interval for the geometric model.}
    \label{fig:geometric}
\end{figure}

As shown by \Cref{fig:editDistanceHistogram}, close carry errors are ``local'' in the sense that they tend to produce small overall edit distance between the model output and the correct answer. This suggests that an error made on one close carry may be relatively independent of other mistakes during the computation. Here we propose a simple one-parameter stochastic model of close carry errors, and show that it often matches the observed patterns of mistakes.

Assume a given addition problem involves $n$ close carries. We imagine that the model proceeds from left to right, producing the correct output digit at each position that does not involve a close carry. Upon reaching a close carry position, the model flips a coin and makes a close carry error with probability $p$. If it makes an error, then the answer will be incorrect. If it does not make an error, it proceeds and, on reaching the next close carry position, flips another independent coin with the same bias $p$. Continuing in this way, the model will ultimately produce a correct answer with probability $(1-p)^n$.

To assess whether this stochastic process is representative of the behavior of real models, we first choose a target of $n=15$ close carries and then select only addition problems with exactly $n$ close carries. (In general, approximately 20\% of columns produce a close carry, so the selected problems have a typical length of $d=75$.) For each model, we set $p$ so that the predicted error matches the observed error rate on these problems, including only problems for which the model either made a close carry error or was correct. We then examine, for each selected problem, the number of \emph{successful} close carries before the first carry error, comparing the distribution observed empirically with the one predicted by the stochastic model. As shown in \Cref{fig:geometric}, the two distributions are often closely (though not perfectly) aligned.

These results are consistent with the idea that some models make close carry errors in a simple, independent way, and do not correlate errors across positions. This is particularly true for Gemma 3 27B. However, not every model's behavior is fully explained in this way. In particular, several of the models show a lower propensity to make a mistake at the final close carry position, which is consistent with a general increase in accuracy for the lowest-order digits.

\section{Discussion}

Recent work has studied the representational power of neural networks (and transformer architectures in particular) \citep{TLAT, Bapo, Mapreduce}. While the specific modeling choices vary, basic arithmetic generally falls in the class of tasks that can be solved exactly using modern architectures. In fact, addition is one of the examples used to explain how to ``think like a transformer''~\citep{TLAT}.

However, our evidence suggests that modern LLMs are not finding reliable implementations for addition, instead resorting to error-prone strategies that give poor results and exhibit design-dependent idiosyncrasies. Bigger models tend to perform better, but are dominated by a simple calculator, especially as the number of digits grows beyond 50. If we expect intelligent systems to perform this basic operation without falling back on specialized tools, then we will need new model designs, data sources, and training strategies.

Of course, arithmetic is itself just an example. Success at arithmetic is easy to measure, and we have ready expectations about how an intelligent system ought to perform. But today's AI systems fail in many unique ways, some of which may be similarly easy to characterize and understand, and others which may be hard to delineate or even see clearly at all. Perhaps LLMs are simply not the right tools for tasks where it is important to identify a single correct answer. Or perhaps even ``soft'' probabilistic reasoning is too much to expect in complex settings. But we will need a clearer understanding of these boundaries and how to move beyond them if we want to develop robust AI deployments that consistently meet our expectations.

\clearpage

\bibliographystyle{plainnat}
\bibliography{bibliography}

\end{document}